\documentclass{bmvc2k}

\usepackage[utf8]{inputenc} 
\usepackage[T1]{fontenc}    
\usepackage{hyperref}       
\usepackage{url}            
\usepackage{booktabs}       
\usepackage{amsfonts}       
\usepackage{nicefrac}       
\usepackage{microtype}      
\usepackage{graphicx}
\usepackage{graphics}
\usepackage{amsmath}
\usepackage{amssymb}
\newcommand{\indep}{\rotatebox[origin=c]{90}{$\models$}}
\usepackage{mathtools}
\usepackage{algorithm}
\usepackage{algpseudocode} 
\usepackage{multirow}
\usepackage{multicol}
\usepackage{tabularx}
\usepackage{boldline}
\usepackage{comment}

\title{Through a fair looking-glass: mitigating bias in image datasets}

\addauthor{Amirarsalan Rajabi}{amirarsalan@knights.ucf.edu}{1}
\addauthor{Mehdi Yazdani-Jahromi}{yazdani@knights.ucf.edu}{2}
\addauthor{Ozlem Ozmen Garibay}{ozlem@ucf.edu}{1,2}
\addauthor{Gita Sukthankar}{gitars@eecs.ucf.edu}{1}

\addinstitution{
 Department of Computer Science\\
 University of Central Florida\\
 Orlando, Florida, USA
}
\addinstitution{
 Department of Industrial Engineering and Management Systems\\
 University of Central Florida\\
 Orlando, Florida, USA
}

\runninghead{Student, Prof, Collaborator}{BMVC Author Guidelines}

\begin{document}
\maketitle
\begin{abstract}
With the recent growth in computer vision applications, the question of how fair and unbiased they are has yet to be explored. There is abundant evidence that the bias present in training data is reflected in the models, or even amplified. Many previous methods for image dataset de-biasing, including models based on augmenting datasets, are computationally expensive to implement. In this study, we present a fast and effective model to de-bias an image dataset through reconstruction and minimizing the statistical dependence between intended variables. Our architecture includes a U-net to reconstruct images, combined with a pre-trained classifier which penalizes the statistical dependence between target attribute and the protected attribute. We evaluate our proposed model on CelebA dataset, compare the results with a state-of-the-art de-biasing method, and show that the model achieves a promising fairness-accuracy combination.
\end{abstract}

\section{Introduction}
\label{sec-intro}

Due to their increased usage within myriad software applications, artificial intelligence algorithms now influence many aspects of people's lives, particularly when they are embedded into decision-support tools used by educators, government agencies, and various industry sectors.
 Thus, it is crucial to make sure that these algorithms are scrutinized to ensure fairness and remove unjust biases. Bias has been shown to exist in several deployed AI systems, including the well known Correlational Offender Management Profiling for Alternative Sanctions (COMPAS). COMPAS is an automated decision making system used by the US criminal justice system for assessing a criminal defendant's likelihood of re-offending. By exploring the risk scores assigned to individuals, this system has been shown to be biased against African Americans \cite{chouldechova2017fair}. Other examples include a version of Google's targeted advertising system in which highly paid jobs were advertised more frequently to men vs.\ women \cite{lambrecht2019algorithmic}.
 
Bias in computer vision is a major problem, often stemming from the training datasets used for computer vision models \cite{tommasi2017deeper}. There is evidence suggesting the existence of multiple types of bias, including capture and selection bias, in popular image datasets \cite{torralba2011unbiased}. The problems arising from bias in computer vision can manifest in different ways. For instance, it is observed that in activity recognition models, when the datasets contain gender bias, the bias is further amplified by the models trained on those datasets \cite{zhao2017men}. Face recognition models may exhibit lower accuracy for some classes of race or gender \cite{buolamwini2018gender}.   

Works such as \cite{wang2020revise,yang2020towards} suggest methods to mitigate bias in visual datasets. Several studies have deployed GANs for bias mitigation in image datasets. For example, \cite{sattigeri2019fairness} modified the value function of GAN to generate fair image datasets. FairFaceGAN~\cite{hwang2020fairfacegan} implements a facial image-to-image translation, preventing unwanted translation in protected attributes. Ramaswamy et al. propose a model to produce training data that is balanced for each protected attribute, by perturbing the latent vector of a GAN \cite{ramaswamy2021fair}. Other studies employing GANs for fair data generation include \cite{choi2020fair, sharmanska2020contrastive}.  

A variety of techniques beyond GANs have been applied to the problems of fairness in AI.  A deep information maximization adaptation network was used to reduce racial bias in face image datasets~\cite{wang2019racial}, and reinforcement learning was used to learn a race-balanced network in \cite{wang2019mitigate}.  Wang et al. propose a generative few-shot cross-domain adaptation algorithm to perform fair cross-domain adaption and improve performance on minority category \cite{wang2021towards}. The work in \cite{xu2021consistent} proposes adding a penalty term into the softmax loss function to mitigate bias and improve fairness performance in face recognition. Quadriento et al. \cite{quadrianto2019discovering} propose a method to discover fair representations of data with the same semantic meaning of the input data. Adversarial learning has also successfully been deployed for this task \cite{zhang2018mitigating, wang2019balanced}.

This paper addresses the issue of a decision-making process being dependent on \emph{protected attributes}, where this dependence should ideally be avoided. From a legal perspective, a protected attribute is an attribute upon which discrimination is illegal  \cite{pessach2020algorithmic}, e.g. gender or race. Let $D = (\mathcal{X},\mathcal{S},\mathcal{Y})$ be a dataset, where $\mathcal{X}$ represents unprotected attributes, $\mathcal{S}$ is the protected attribute, and $\mathcal{Y}$ be the target attribute. If in the dataset $D$, the target attribute is not independent of the protected attribute ($\mathcal{Y} \not\perp \mathcal{S}$), then it is very likely that the decisions $\mathcal{\hat{Y}}$ made by a decision-making system which is trained on $D$, is also not independent of the protected attribute ($\mathcal{\hat{Y}} \not\perp \mathcal{S}$).

We propose a model to reconstruct an image dataset to reduce statistical dependency between a protected attribute and target attribute. We modify a U-net \cite{ronneberger2015u} to reconstruct the image dataset and apply the Hilbert-Schmidt norm of the cross-covariance operator \cite{gretton2005measuring} between reproducing kernel Hilbert spaces of the target attribute and the protected attribute, as a measure of statistical dependence. Unlike many previous algorithms, our proposed method doesn't require training new classifiers on the unbiased data, but instead reconstructing images in a way that reduces the bias entailed by using the same classifiers. 

In Section~\ref{sec-method} we present the problem, the notion of independence, and our proposed methodology. In Section~\ref{sec-experiments} we describe the CelebA dataset and the choice of feature categorization, introduce the baseline model with which we compare our results \cite{ramaswamy2021fair}, our model's implementation details, and finally present the experiments and results.

Bias mitigation methods can be divided into three general categories of \emph{pre-process}, \emph{in-process}, and \emph{post-process}. Pre-process methods include modifying the training dataset before feeding it to the machine learning model. In-process methods include adding regularizing terms to penalize some representation of bias during the training process. Finally, post-process methods include modifying the final decisions of the classifiers \cite{hardt2016equality}. Kamiran and Calders~ \cite{kamiran2012data} propose methods such as suppression which includes removing attributes highly correlated with the protected attribute, reweighing, i.e. assigning weights to different instances in the data, and massaging the data to change labels of some objects. Bias mitigation methods often come at the expense of losing some accuracy, and these preliminary methods usually entail higher fairness-utility cost. More sophisticated methods with better results include using generative models to augment the biased training dataset with unbiased data \cite{ramaswamy2021fair}, or training the models on entirely synthetic unbiased data \cite{rajabi2021tabfairgan}.
Wang et al.\cite{wang2020towards} provide a set of analyses and a benchmark to evaluate and compare bias mitigation techniques in visual recognition models.

\section{Methodology}
\label{sec-method}
Consider a dataset $D = (\mathcal{X},\mathcal{S},\mathcal{Y})$, where $\mathcal{X}$ is the set of images, $\mathcal{Y} = \{+1, -1\}$ is the target attribute such as attractiveness, and $\mathcal{S} = \{A,B,C,...\}$ is the protected attribute such as gender. Assume there exists a classifier $f:(\mathcal{X}) \rightarrow \mathcal{Y}$, such that the classifier's prediction for target attribute is not independent from the protected attribute, i.e. $f(\mathcal{X}) \not\perp \mathcal{S}$. Our objective is to design a transformation $g:\mathcal{X} \rightarrow \widetilde{\mathcal{X}}$, such that 1) $f(\widetilde{\mathcal{X}}) \perp \mathcal{S}$, i.e. the classifier's predictions for target attribute is independent of the protected attribute , and 2) $f(\widetilde{\mathcal{X}}) \approx f(\mathcal{X})$, i.e. the classifier still achieves high accuracy. 

In  other words we want to train a network to transform our original images, such that if the classifiers that are trained on the original and unmodified images, are used to predict the target attribute (attractiveness in our example) from the transformed version of an image, they still achieve high accuracy, while the predictions of those classifiers are independent of the protected attribute (gender in our example). It should be noted that we are not seeking to train new classifiers, but rather only aim to modify the input images. This is a main distinction between our methodology and most of other techniques (e.g. \cite{quadrianto2019discovering} and \cite{ramaswamy2021fair}), in which the process includes training new classifiers on modified new image datasets and achieving \emph{fair classifiers}. 

Our proposed model consists of a U-net \cite{ronneberger2015u} as the neural network that transforms the original images. This type of network was originally proposed for medical image segmentation, and has been widely used since its introduction. The encoder-decoder network consists of two paths, a contracting path consisting of convolution and max pooling layers, and a consecutive expansive path consisting of upsampling of the feature map and convolutions. Contrary to \cite{ronneberger2015u} where each image is provided with a segmented image label, we provide our U-net with the exact same image as the label, and alter the loss function from cross-entropy to mean squared error, so that the network gets trained to produce an image as close to the original image as possible, in a pixel-wise manner. 

\begin{figure}[t!]
\centering
	\includegraphics[width=0.7\linewidth,keepaspectratio]{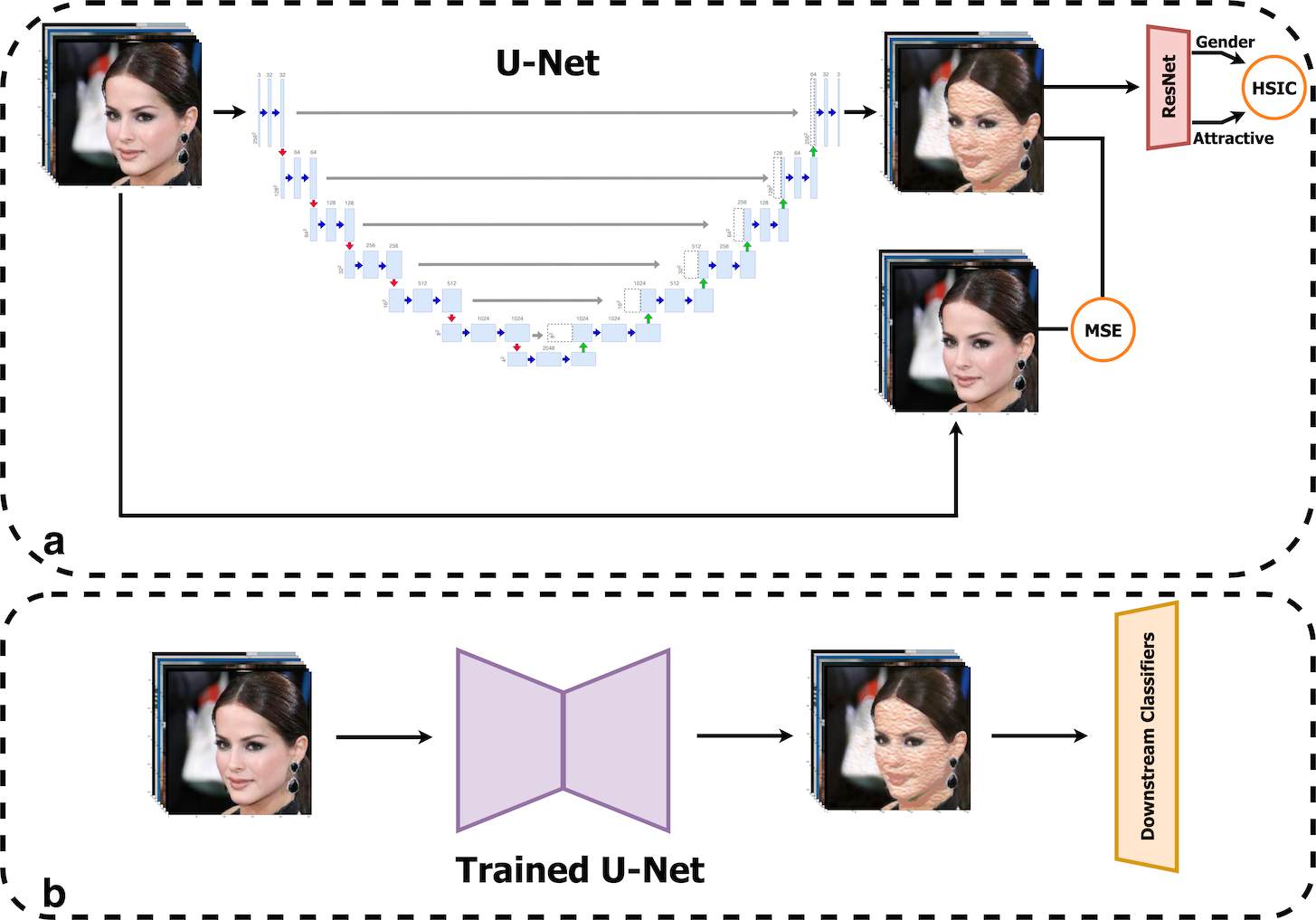}
	\caption{Our model consists of an encoder-decoder (U-net) and a double-output pre-trained ResNet classifier. First, the output batch of the U-net (reconstructed images) is compared with the original batch of images by calculating MSE loss. Then, the output batch of the U-net passes through the ResNet and statistical dependency of the two vectors is calculated by HSIC. Detailed architecture of the U-net is described in the supplementary material.}
	\label{fig-network}
\end{figure}

While some previous fairness studies consider \textit{decorrelating} the target attribute from the protected attributes, what must be ultimately sought however, is independence between the protected attribute and the target attribute. Dealing with two random variables which are uncorrelated is easier than independence, as two random variables might have a zero correlation, and still be dependent (e.g. two random variables $A$ and $B$ with recordings $A=[-2,-1,0,1,2]$ and $B=[4,1,0,1,4]$ have zero covariance, but are apparently not independent).  Given a Borel probability distribution $\mathbf{P}_{ab}$ defined on a domain $\mathcal{A} \times \mathcal{B}$, and respective marginal distributions $\mathbf{P}_a$ and $\mathbf{P}_b$ on $\mathcal{A}$ and $\mathcal{B}$, independence of $a$ and $b$ ($a \indep b$) is equal to $\mathbf{P}_{xy}$ factorizing as $\mathbf{P}_x$ and $\mathbf{P}_y$. Furthermore, two random variables $a$ and $b$ are independent, if and only if any bounded continuous function of the two random variables are uncorrelated \cite{gretton2005kernel}. 

Let $\mathcal{F}$ and $\mathcal{G}$ denote all real-value  functions defined on domains $\mathcal{A}$ and $\mathcal{B}$ respectively. In their paper Gretton et al. \cite{gretton2005measuring} define the Hilbert-Schmidt norm of the cross-covariance operator:
\begin{equation}
    HSIC(\mathbf{P}_{ab}, \mathcal{F}, \mathcal{G}) \coloneqq ||C_{ab}||^2_{HS}
    \label{eq-HSIC-def}
\end{equation}

\noindent where $C_{ab}$ is the cross-covariance operator. They show that if $||C_{ab}||^2_{HS}$ is zero, then $cov(f,g)$ will be zero for any $f \in \mathcal{F}$ and $g \in \mathcal{G}$, and therefore the random variables $a$ and $b$ will be independent. Furthermore, they show if $\mathcal{Z} \coloneqq {(a_1, b_1),...,(a_n,b_n)} \in \mathcal{A} \times \mathcal{B}$ are a series of n independent observations drawn from $\mathbf{P}_{ab}$, then a (biased) estimator of \textbf{HSIC} is \cite{gretton2005measuring}:

\begin{equation}
    HSIC(\mathcal{Z},\mathcal{F}, \mathcal{G}) \coloneqq (n-1)^{-2}\mathbf{tr}(KHLH)
    \label{eq-HSIC-est}
\end{equation}

\noindent where $H,K,L \in \mathbb{R}^{n \times n}$, $K$ and $L$ are Gram matrices \cite{horn2012matrix}, $K_{ij} \coloneqq k(a_i, a_j)$, $L_{ij} \coloneqq l(b_i, b_j)$, $k$ and $l$ are universal kernels, and $H_{ij} \coloneqq \delta_{ij} - n^{-1}$ centers the observations in feature space. We use Hilbert-Schmidt independence criteria to penalize the model for dependence between the target attribute and the protected attribute.


\begin{figure*}
\begin{center}
\includegraphics[width=0.8\linewidth,keepaspectratio]{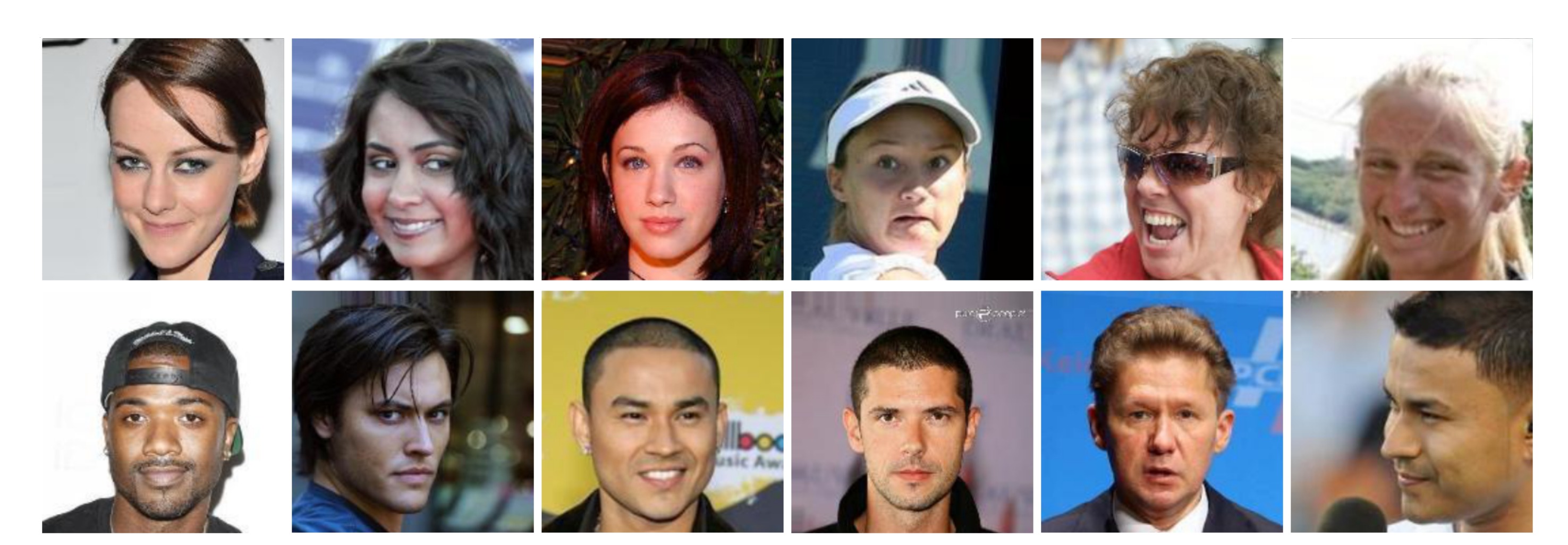}
	\caption{Examples of CelebA dataset original images. Images in the first row are labeled \texttt{not Male} and images in the second row are labeled \texttt{Male}. In each row, the first three images are labeled \texttt{Attractive} and the last three images are labeled \texttt{not Attractive}.}
	\label{fig-faces-explanatory}    
\end{center}
\end{figure*}


\subsection{Training Loss Function}
\label{sec:training}
We seek to modify a set of images, such that 1) the produced images are close to the original images, and 2) the predicted target attribute is independent from the predicted protected attribute. In the optimization problem, image quality (1) is measured by pixel-wise MSE loss. For independence (2), consider our U-net network as a mapping from original image to the transformed image, i.e. $U_w(\mathbf{x}) = \widetilde{\mathbf{x}}$. Consider also a function $h:\mathcal{X} \rightarrow [0,1]\times[0,1]$, where $h(\mathbf{x}_i) = (h_1(\mathbf{x}_i), h_2(\mathbf{x}_i)) = (\mathrm{P}(y_i=1|\mathbf{x}_i), \mathrm{P}(s_i=1|\mathbf{x}_i))$. Our objective is to train the parameters of $U_w$ such that $h_1(U_w(\mathbf{x})) \indep h_2(U_w(\mathbf{x}))$, i.e. $h_1(U_w(\mathbf{x}))$ is independent of $h_2(U_w(\mathbf{x}))$ . 

Given $X$ representing a batch of N training images and $\widetilde{X}$ representing the transformed batch, our formal optimization problem is as follows:

\begin{equation}
\begin{split}
    \mathop{\mathrm{minimize}}_{U_w} &\underbrace{\frac{1}{NCWH} \sum_{n=1}^{N} \sum_{i,j,k} (\mathbf{x}_{ijk}^{n} - \widetilde{\mathbf{x}}_{ijk}^{n})^2}_\textrm{image accuracy} \\ 
    &+ \lambda \times \; \underbrace{HSIC(h_1(\widetilde{X}), h_2(\widetilde{X}))}_\textrm{independence}
\end{split}
\end{equation}

\noindent where $N$ is the number of samples, $C$ is the number of channels of an image, $W$ is the width of an image, $H$ is the height of an image, and $\lambda$ is the parameter that controls the trade-off between accuracy of the transformed images and independence (fairness). In practice, the mapping function $U_w$ that we use is a U-net, the function $h(\cdot)$ is a pre-trained classifier with two outputs $h_1$ and $h_2$, each being the output of a Sigmoid function within the range of $[0,1]$, where $h_1 = \mathrm{P}(Y = 1|X)$ (a vector of size $N$), and $h_2=\mathrm{P}(S = 1|X)$ (also a vector of size $N$), and $HSIC(\cdot,\cdot)$ denotes Hilbert-Schmidt Independence Criteria. 

Figure~\ref{fig-network} shows the network architecture and a schematic of the training procedure. Consider a batch of original images $X$ entering the U-net. The U-net then produces the reconstructed images $U_w(X) = \widetilde{X}$. To calculate the \emph{image accuracy} part of the loss function, the original image batch $X$ is provided as label and the Mean Squared Error is calculated to measure the accuracy of the reconstructed images. The ResNet component in Figure~\ref{fig-network} is our $h(\cdot)$ function as described before, which is a pre-trained ResNet classifier that takes as input a batch of images and returns two probability vectors. The second part of the loss function, \emph{independence}, is calculated by entering the reconstructed images $\widetilde{X}$ into this ResNet classifier, and calculating the HSIC between the two vectors. 

As noted before, the image dataset is reconstructed in a way that using them on the original biased classifiers, will result in an improvement in classifications. This is dissimilar to some previous works such as \cite{ramaswamy2021fair} and \cite{quadrianto2019discovering}, in which the model training process includes augmenting the original dataset with generated images and training new fair classifiers \cite{ramaswamy2021fair}, or discovering fair representations of images and subsequently training new classifiers \cite{quadrianto2019discovering}.

\section{Experiments}
\label{sec-experiments}
In this section, we test the methodology described in Section~\ref{sec-method} on CelebA dataset \cite{liu2015faceattributes}. We first introduce the CelebA dataset and the attribute categories in CelebA. We then describe the implementation details of our model. Subsequently, the method described in Ramaswamy et al. \cite{ramaswamy2021fair} and the two versions of it that we use as baseline models to compare our results with are introduced. Finally, we introduce evaluation metrics and present the results.

\subsection{CelebA dataset}
\label{sec-celeba}
CelebA is a popular dataset that is widely used for training and testing models for face detection, particularly recognising facial attributes. It consists of 202,599 face images of celebrities, with 10,177 identities. Each image is annotated with 40 different binary attributes describing the image, including attributes such as \texttt{Black\_Hair}, \texttt{Pale\_Skin}, \texttt{Wavy\_Hair}, \texttt{Oval\_Face}, \texttt{Pointy\_Nose}, and other attributes such as \texttt{Male}, \texttt{Attractive}, \texttt{Smiling}, etc. The CelebA dataset is reported to be biased \cite{zhang2018examining}. In this experiment, we consider \texttt{Male} attribute as the protected attribute (with $\texttt{Male} = 0$ showing the image does not belong to a man and $\texttt{Male} = 1$ showing the image belongs to a man), and \texttt{Attractive} to be the target attribute. We divide the dataset into train and test sets, with train set containing 182,599 and test set containing 20,000 images. In the training set, $67.91\%$ of images with $\texttt{Male}=0$ are annotated to be attractive ($\texttt{Attractive}=1$), while only $27.93\%$ of images with $\texttt{Male}=1$ are annotated as being attractive ($\texttt{Attractive}=1$). This shows bias exists against images with $\texttt{Male}=1$.

In order to compare our results with \cite{ramaswamy2021fair}, we follow their categorization of CelebA attributes. Leaving out gender (\texttt{Male}) as the protected attribute, among the rest 39 attributes in CelebA dataset, \cite{ramaswamy2021fair} eliminates some attributes such as \texttt{Blurry} and \texttt{Bald} as they contain less than 5\% positive images. The remaining 26 attributes is subsequently categorized into three groups. \emph{inconsistently-labeled} attributes are the ones that by visually examining sets of examples, the authors often disagree with the labeling and could not distinguish between positive and negative examples \cite{ramaswamy2021fair}. This group includes attributes such as \texttt{Straight\_Hair}, and \texttt{Big\_Hair}. The second group of attributes are the ones that are called \emph{gender-dependent} and the images are labeled to have (or not have) attributes based on the perceived gender \cite{ramaswamy2021fair}. These include attributes such as \texttt{Young, Arched\_Eyebrows} and \texttt{Receding\_Hairline}. Finally, the last group of attributes are called \emph{gender-independent}. These attributes are fairly consistently labeled and are not much dependent on gender expression. This group includes attributes such as \texttt{Black\_Hair, Bangs}, and \texttt{Wearing\_Hat}. The list of all attributes is provided in supplementary material. 

In order to compare our results with \cite{ramaswamy2021fair}, we follow their categorization of CelebA attributes. Leaving out gender (\texttt{Male}) as the protected attribute, among the rest 39 attributes in CelebA dataset, \cite{ramaswamy2021fair} eliminates some attributes such as \texttt{Blurry} and \texttt{Bald} as they contain less than 5\% positive images. The remaining 26 attributes is subsequently categorized into three groups. \emph{inconsistently-labeled} attributes are the ones that by visually examining sets of examples, the authors often disagree with the labeling and could not distinguish between positive and negative examples \cite{ramaswamy2021fair}. This group includes \texttt{Straight\_Hair, Big\_Lips, Big\_Nose, Oval\_Face, Pale\_Skin}, and \texttt{Wavy\_Hair}. The second group of attributes are the ones that are called \emph{gender-dependent} and the images are labeled to have (or not have) attributes based on the perceived gender \cite{ramaswamy2021fair}. These include \texttt{Young, Arched\_Eyebrows, Attractive, Bushy\_Eyebrows, Pointy\_Nose}, and \texttt{Receding\_Hairline}. Finally, the last group of attributes are called \emph{gender-independent}. These attributes are fairly consistently labeled and are not much dependent on gender expression. This group of attributes include \texttt{Black\_Hair, Bangs, Blond\_Hair, Brown\_Hair, Chubby, Wearing\_Earrings, Bags\_Under\_Eyes, Eyeglasses, Gray\_Hair, High\_Cheekbones, Mouth\_Slightly\_Open, Narrow\_Eyes, Smiling}, and \texttt{Wearing\_Hat}.

\begin{figure*}
\centering
\includegraphics[width=0.8\linewidth,keepaspectratio]{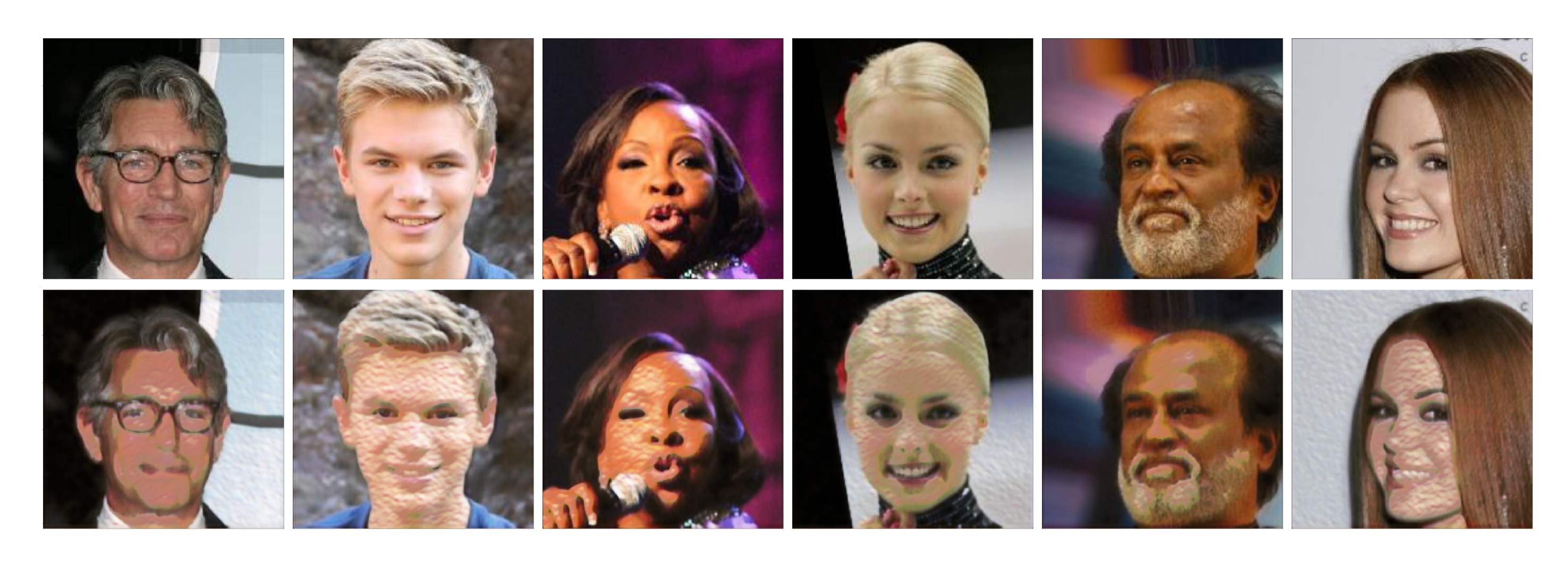}
	\caption{Examples of CelebA dataset images and how the model reconstructs them. The first row shows a set of images from the original testing set, and the second row shows the reconstructed images.}
\label{fig-faces}
\end{figure*}

\subsection{Attribute classifiers}
For attribute classifiers, we use ResNet-18 pre-trained on ImageNet, in which the last layer is replaced with a layer of size one, along with a Sigmoid activation for binary classification. We train all models for 5 epochs with batch sizes of 128. We use the Stochastic Gradient Descent optimizer with a learning rate of 1e-3 and momentum of 0.9. We use a step learning rate decay with step size of 7 and factor of 0.1. After training, we will have 26 classifiers that receive an image and perform a binary classification on their respective attribute. 

\subsection{Implementation details}
\label{sec-implementationdetails}
As shown in Figure~\ref{fig-network}, a ResNet-18 network is used to accompany the U-net to produce predictions for \texttt{Male} and \texttt{Attractive}. Prior to training the U-net, the ResNet-18 \cite{russakovsky2015imagenet} which is pre-trained on ImageNet, is modified by replacing its output layer with a layer of size two, outputing the probability of attractiveness and gender. The ResNet-18 is then trained for 5 epochs on the train set, with a batch size of 128. We use the Stochastic Gradient Descent optimizer with a learning rate of 1e-3 and momentum of 0.9. We use a step learning rate decay with step size of 7 and factor of 0.1. After the ResNet is trained and prepared, we train the U-net as described in Section~\ref{sec-method} on the train set. The detailed architecture of the U-net is described in Supplementary Material. In our implementation of biased estimator of HSIC estimator in Equation~\ref{eq-HSIC-est}, we use Gaussian RBF kernel function for $k(\cdot,\cdot)$ and $l(\cdot,\cdot)$. The training was conducted on a machine with two NVIDIA GeForce RTX 3090, and each training of the U-Net took 1 hour. When the training is complete, the U-net is ready to reconstruct images. Figure~\ref{fig-faces} shows six examples of how the U-net modifies the original images. We train our model for 5 epochs with an $\lambda = 0.07$.

\subsection{Comparison with baseline models}
We compare our results with Ramaswamy et al.'s method, described in their paper `Fair Attribute Classification through Latent Space De-biasing' \cite{ramaswamy2021fair}. Building on work by \cite{denton2019image} which demonstrates a method to learn interpretable image modification directions, they develop an improved method by perturbing latent vector of a GAN, to produce training data that is balanced for each protected attribute. By augmenting the original dataset with the generated data, they train target classifiers on the augmented dataset, and show that these classifiers will be fair, with high accuracy. The second model that we compare our results with is explicit removal of biases from neural network embeddings, presented in \cite{alvi2018turning}. The authors provide an algorithm to remove multiple sources of variation from the feature representation of a network. This is achieved by including secondary branches in a neural network with the aim to minimize a confusion loss, which in turn seeks to change the feature representation of data such that it becomes invariant to the spurious variations that are desired to be removed. 

We implement Ramaswamy et al.'s method as follows: As mentioned in their paper, we used progressive GAN with 512-D latent space trained on the CelebA training set from the PyTorch GAN Zoo. We use 10,000 synthetic images and label the synthetic images with a ResNet-18 (modified by adding a fully connected layer with 1,000 neurons). Then we trained a linear SVM to learn the hyper-planes in the latent space as proposed in the original paper. We generate $\mathcal{X}_{syn}$ (160,000 images) to generate a synthetic dataset which aims to de-bias \texttt{Male} from all 26 attributes one by one. Next, we train ResNet-18 classifiers on the new datasets consisting of augmenting $\mathcal{X}$ and $\mathcal{X}_{syn}$. We call this model as \emph{GANDeb}. We use the implementation of \cite{alvi2018turning} with the uniform confusion loss $-(1/\vert D \vert) \sum_{d}{\log q_d}$ provided in \cite{wang2020towards}.

\subsection{Evaluation metrics}
In evaluating the results of our model with the baseline models, three metrics are used. To capture the accuracy of the classifiers, we measure the \emph{average precision}. This metric combines precision and recall at every position and computes the average. A higher average precision (\textbf{AP}) is desired. To measure fairness, there are multiple metrics proposed in the literature \cite{mehrabi2021survey}. Among the most commonly used metrics is \emph{demographic parity} (\textbf{DP}). This metric captures the disparity of receiving a positive decision among different protected groups ($|P(\hat{Y}=1|S=0) - P(\hat{Y}=1|S=1)|$). A smaller \textbf{DP} shows a fairer classification and is desired. Finally for our last fairness measure, we follow \cite{lokhande2020fairalm} and \cite{ramaswamy2021fair} and use \emph{difference in equality of opportunity} (\textbf{DEO}), i.e. the absolute difference between the true positive rates for both gender expressions ($|TPR(S=0) - TPR(S=1)|$). A smaller \textbf{DEO} is desired. 

\subsection{Results}
\label{sec-results}
All the values reported in this section, are evaluated on the same test set. Prior to comparing the results of our method with the comparison models, to assess the original training data, the performance of baseline classifiers being trained on the original train set, and tested on the test set is presented. The AP, DP, and DEO values of classifiers trained on the original training set is shown in Table~\ref{tab-results} under \emph{Baseline}. Looking into Baseline values, the AP of classifiers for gender-independent category of attributes is higher than gender-dependent category, and the AP of inconsistent category is less than the other two categories. As expected, DP and DEO for gender-dependent category of attributes is higher than the other two categories.

In Table~\ref{tab-results}, we compare our model with GAN Debiasing (GanDeb) \cite{ramaswamy2021fair}, Adversarial debiasing (AdvDb) presented in \cite{alvi2018turning}, and the Baseline on the original data. Looking into the average precision scores, the results show that GanDeb is slightly performing better than Ours. This is anticipated, since half of the training data for GanDeb consists of the original images, and therefore a higher average precision is expected. AdvDb on the other hand is performing poorly in terms of average precision, with average precision scores far away from other models. 

Looking into demographic parity scores, the results show that GanDeb falls behind the other two models in two out of three attribute categories. While Ours is performing better for gender dependent and gender independent attribute categories. Looking into the third fairness measure, difference in equality of opportunity, AdvDb and ours are performing better than GanDeb in all three categories of attributes. Ours beats AdvDb for inconsistent attributes category, AdvDb beats Ours in gender dependent category, and AdvDb slightly beats Ours for gender independent category of attributes. In summary, Ours is close to GanDeb in terms of maintaining high average precision scores, which means higher accuracy of prediction, while beating GanDeb in terms of fairness metrics. Also, while AdvDb performance in terms of fairness enforcement is better than ours in 3 out of 6 cases, it falls behind significantly in terms of average precision. 

\begin{figure*}[t!]
\centering
    \includegraphics[width=1.0\linewidth,keepaspectratio]{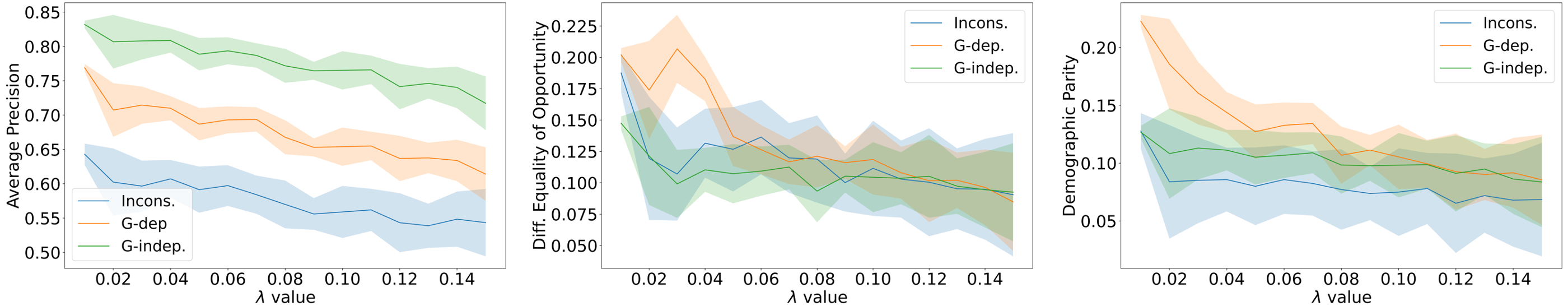}
	\caption{Exploring the trade-off between accuracy and fairness by incremental increasing of parameter $\lambda$. Each data point is the average over three trainings, with standard deviation of the three trainings shown as confidence intervals.}
\label{fig-ablation}
\end{figure*}

To explore the trade-off between fairness and precision, we perform the following experiment: $\lambda$ was increased between $[0.01, 0.15]$ in steps of 0.01, and for each value of $\lambda$, the model was trained three times, each time for 1 epoch. Figure~\ref{fig-ablation} shows how AP, DEO, and DP change. The results show that by increasing $\lambda$, precision decreases while fairness measures improve.

\begin{table*}[ht]
\resizebox{\linewidth}{!}{%
\begin{tabular}{cl|cccc|cccc|cccc|}
\cline{3-11}
\multicolumn{2}{c|}{\textbf{}} & \multicolumn{3}{c|}{$\textbf{AP} \uparrow$}                                                   & \multicolumn{3}{c|}{$\textbf{DP} \downarrow$}                                                     & \multicolumn{3}{c|}{$\textbf{DEO} \downarrow$}                                            \\ \cline{3-11} 
\multicolumn{2}{c|}{}          & \multicolumn{1}{c|}{Incons.} & \multicolumn{1}{c|}{G-dep}    & \multicolumn{1}{c|}{G-indep} & \multicolumn{1}{c|}{Incons.} & \multicolumn{1}{c|}{G-dep}        & \multicolumn{1}{c|}{G-indep}  & \multicolumn{1}{c|}{Incons.} & \multicolumn{1}{c|}{G-dep}  & \multicolumn{1}{c|}{G-indep}   \\ \hline
\multicolumn{2}{|c|}{Baseline}  & \multicolumn{1}{c|}{0.667}    & \multicolumn{1}{c|}{0.79}  & \multicolumn{1}{c|}{0.843}        & \multicolumn{1}{c|}{0.147}    & \multicolumn{1}{c|}{0.255}    & \multicolumn{1}{c|}{0.137} & \multicolumn{1}{c|}{0.186}    & \multicolumn{1}{c|}{0.243} &  \multicolumn{1}{c|}{0.163} \\ \hline
\multicolumn{2}{|c|}{GanDeb}   & \multicolumn{1}{c|}{0.641}     & \multicolumn{1}{c|}{0.763}  & \multicolumn{1}{c|}{0.831}       & \multicolumn{1}{c|}{0.106}    & \multicolumn{1}{c|}{0.233}       & \multicolumn{1}{c|}{0.119} & \multicolumn{1}{c|}{0.158}    & \multicolumn{1}{c|}{0.24}  & \multicolumn{1}{c|}{0.142} \\ \hline
\multicolumn{2}{|c|}{AdvDb}   & \multicolumn{1}{c|}{0.243}     & \multicolumn{1}{c|}{0.333}  & \multicolumn{1}{c|}{0.218}       & \multicolumn{1}{c|}{0.091}    & \multicolumn{1}{c|}{0.169}       & \multicolumn{1}{c|}{0.121} & \multicolumn{1}{c|}{0.136}    & \multicolumn{1}{c|}{0.149}  & \multicolumn{1}{c|}{0.098} \\ \hline
\multicolumn{2}{|c|}{Ours} & \multicolumn{1}{c|}{0.618}    & \multicolumn{1}{c|}{0.732} & \multicolumn{1}{c|}{0.839}        & \multicolumn{1}{c|}{0.097}    & \multicolumn{1}{c|}{0.146} & \multicolumn{1}{c|}{0.118}          & \multicolumn{1}{c|}{0.124}    & \multicolumn{1}{c|}{0.172} & \multicolumn{1}{c|}{0.114} \\ \hline
\end{tabular}
}
\caption{Comparing the results of our model with Baseline, GAN debiasing (GanDeb), and Adversarial debiasing (AdvDb). Showing AP (Average Precision, higher the better), DP (Demographic Parity, lower the better), and DEO (Difference in Equality of Opportunity, lower the better) values for each attribute category. Each number is the average over all attributes within that specific attribute category.}
\label{tab-results}
\end{table*}

\subsection{Interpretation and the effect on other attributes}
\begin{figure*}[t!]
\centering
    \includegraphics[width=0.6\linewidth,keepaspectratio]{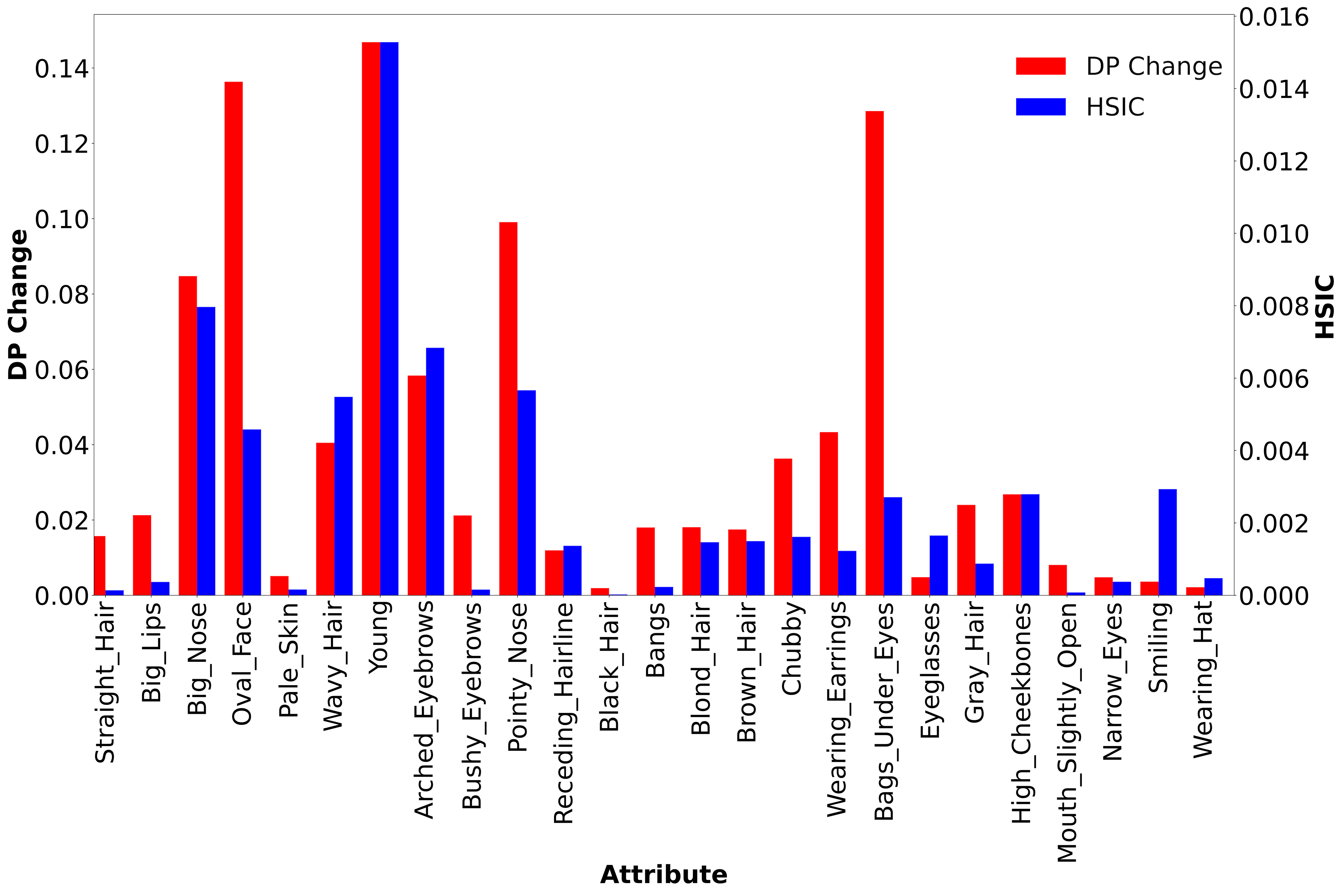}
	\caption{Displaying the relationship between an attribute's statistical dependence on \texttt{Attractive} attribute, and the extent to which the model modifies that attribute. Blue bars show the HSIC between each attribute with \texttt{Attractive} attribute in the original data. Red bars show the absolute difference in demographic parity of each attribute's classifier, acting on original images and transformed images, respectively.}
\label{fig-attributes}
\end{figure*}

In this section, we aim to display the correspondence between an attribute's relationship with \texttt{Attractive} attribute, and the extent to which the model modifies that attribute. To do so, for each attribute, we record two values, namely HSIC value between that attribute and the \texttt{Attractive} attribute, and the change in demographic parity. To calculate the change in demographic parity, we first calculate the demographic parity of the classifier for that specific attribute, when the classifier classifies the original testing set images (similar to \emph{Baseline} in previous tables, but for each attribute separately). We then calculate the demographic parity of the classifier for that specific attribute, when the classifier receives the modified training images \textbf{Ours(5,0.07)}. We then subtract the two values, to get the change in demographic parity for that specific attribute. Figure~\ref{fig-attributes} presents the results, with the red bars showing the change in demographic parity for each attribute, and the blue bars showing the statistical dependence measured by HSIC, between each attribute with \texttt{Attractive} attribute, in the original training data. The results show that the absolute change in demographic parity is positively correlated with that attribute's statistical dependence with the attribute \texttt{Attractive}, with a Pearson correlation coefficient of 0.757. For instance, we observe large changes in demographic parity for attributes such as \texttt{Young, Big\_Nose, Pointy\_Nose, Oval\_Face}, and \texttt{Arched\_Eyebrows}, as they are typically associated with being attractive, and therefore reflected in the CelebA dataset labels. 

\section{Conclusions}
\label{sec-conclusions}
We proposed an image reconstruction process to mitigate bias against a protected attribute. The model's performance was evaluated on CelebA dataset and compared with an augmentation based method developed by \cite{ramaswamy2021fair}. The proposed model showed promising results in mitigating bias while maintaining high precision for classifiers. An interesting aspect of the results is that although we only explicitly train the U-net to remove dependence between the target attribute (\texttt{Attractive}) and the protected attribute (\texttt{Male}), classifiers related to many other attributes, most of which have a statistical dependency with the target attribute, become `fairer'. An advantage of the proposed model is that it does not rely on modifying downstream classifiers, and rather includes only modifying the input data, hence making it suitable to be deployed in an automated machine learning pipeline more easily and with lower cost. As a potential future direction, we intend to consider the problem in a situation where multiple protected attributes are present, and attributes are non-binary. We also intend to apply similar methodology on other data types such as tabular data.

\bibliography{egbib}
\end{document}